\documentclass[conference]{IEEEtran}
\usepackage{blindtext, graphicx}
\ifCLASSINFOpdf
\else
\fi
\begin{document}
%
\title{Araguaia Medical Vision Lab at ISIC 2017 Skin Lesion Classification Challenge}

\author{\IEEEauthorblockN{Rafael Teixeira Sousa}
\IEEEauthorblockA{Universidade Federal\\de Mato Grosso\\
Email: rafaelsousa@ufmt.br}
\and
\IEEEauthorblockN{Larissa Vasconcellos de Moraes}
\IEEEauthorblockA{Universidade Federal\\de Mato Grosso}
}


%


\maketitle

\begin{abstract}
This paper describes the participation of Araguaia Medical Vision Lab at the International Skin Imaging Collaboration 2017 Skin Lesion Challenge. We describe the use of deep convolutional neural networks in attempt to classify images of Melanoma and Seborrheic Keratosis lesions. With use of finetuned GoogleNet and AlexNet we attained results of $0.950$ and $0.846$ AUC on Seborrheic Keratosis and Melanoma respectively.
\end{abstract}

\begin{IEEEkeywords}
IEEEtran, journal, \LaTeX, paper, template.
\end{IEEEkeywords}

%
\IEEEpeerreviewmaketitle

\section{Motivation and Overview}
This paper is a small overview of Araguaia Medical Vision Lab (AMVL) at the International Skin Imaging Collaboration (ISIC) 2017 challenge, more specifically the skin lesion classification task. Our main objective is to perform an automatic classification of skin lesions on two main tasks, the Melanoma and Seborrheic Keratosis recognition, using the image dataset available by ISIC, which was already diagnosed by specialists and used as ground truth.
The algorithm proposed a combination of deep convolutional neural networks (CNN), GoogleNet\cite{DBLP:journals/corr/SzegedyLJSRAEVR14} and AlexNet\cite{krizhevsky2012imagenet}, fine-tuned \cite{shie2015transfer} with augmented skin lesion images. The next sessions will describe how was the training and evaluation process.

\section{Image Pre-processing}
The original dataset is composed by 2000 images, with 374 samples of Melanoma and 254 samples of Seborrheic Keratosis, the other 1372 are defined as Nevus. The images have different sizes from 1022 x 767 to 6748 x 4499.

The first step was split a validation set with around 20\% of images from each class to evaluate the neural network performance during the training stage.

All train dataset pass through a pre-process filter which applied random shear, zoom, and vertical and horizontal shift and flip. This step was necessary to increase the dataset size (around 5 times), make it less unbalanced and improve the neural network accuracy.

\section{Network Training and Evaluation}

To perform all the training and evaluation we used Nvidia Digits interface running with Caffe \cite{jia2014caffe}. 

\subsection{Seborrheic Keratosis Task}

On Seborrheic classification task we got the best result resizing all training images to 350 x 350 without losing proportions and training an AlexNet pre-trained on ImageNet classification task dataset by the Berkeley Vision and Learning Center (ref). The network was trained over 30 epochs with a Stochastic Gradient Descent (SGD) using three stages, first 10 epochs with learning rate of $0.001$, than 10 epochs with $0.0001$, and 10 more with $0.00001$.
The top accuracy on validation set was $98.2\%$, with $97.73\%$ on Seborrheic class and $98.28\%$ on non-Seborrheic. On validation dataset from ISIC the network we got $89.3\%$ accuracy and $0.950$ of Area under Roc Curve (AUC), with sensibility of $0.786$ and specificity of $0.935$.

\subsection{Melanoma Task}

On Melanoma task we got our best result doing an average between three networks, GoogleNet 256, GoogleNet 224 and AlexNet 224, where all was pre-trained on ImageNet classification dataset.

GoogleNet 256 was retrained over 72 epochs with 256 x 256 resized images with random crop to 224 x 224, which is the network standard size, the learning was with SGD and three learning rate steps $0.001$, $0.0001$, $0.00001$, the best result was got with the last epoch.

GoogleNet 224 was retrained over 50 epochs with images resized to 224 x 224 without the random crop with SGD and learning steps of $0.005$, $0.0025$, $0.00125$, $0.000625$ and $0.0003125$. Best result was at epoch 15

AlexNet 224 was retrained over 30 epochs with 224 x 224 images, SGD and three learning rate steps $0.001$, $0.0001$, $0.00001$ over 30 epochs, best result was at epoch 15.

The result was obtained by the arithmetic mean between the three networks softmax output, making it more balanced between classes. On validation dataset from ISIC the network we got $84.7\%$ accuracy and $0.846$ of AUC, with sensibility of $0.633$ and specificity of $0.900$.

\section{Conclusion}

The CNNs got great results on both tasks, but seem to lack on sensibility, probably because the unbalanced dataset with few positive images, making it harder to generalize the visual features of lesions.



\bibliographystyle{IEEEtran}
%
\bibliography{main}

%




\end{document}